\documentclass[letterpaper]{article}
\pdfoutput=1
\usepackage{natbib}
\usepackage{alifeconf}
\usepackage{color}
\usepackage{refcount}
\title{Designing a minimalist socially aware robotic agent for the home}
\author{Matthew R. Francisco$^{1}$, Ian Wood$^{1}$, Selma \v{S}abanovi\'{c}$^{1,2}$, \and Luis M. Rocha$^{1,2,3}$ \\
\mbox{}\\
$^1$School of Informatics and Computing, Indiana University, Bloomington, IN 47406, USA\\
$^2$Cognitive Science Program, Indiana University, Bloomington, IN 47406, USA \\
$^{3}$Instituto Gulbenkian de Ciencia, Oeiras, Portugal \\
francm@indiana.edu}

\begin{document}
\maketitle

\begin{abstract}
We present a minimalist social robot that relies on long time-series of low resolution data such as mechanical vibration, temperature, lighting, sounds and collisions.  Our goal is to develop an experimental system for growing socially situated robotic agents whose behavioral repertoire is subsumed by the social order of the space.   To get there we are designing robots that use their simple sensors and motion feedback routines to recognize different classes of human activity and then associate to each class a range of appropriate behaviors. We use the \emph{Katie Family} of robots, built on the \emph{iRobot Create} platform, an \emph{Arduino Uno}, and a \emph{Raspberry Pi}.
We describe its sensor abilities and exploratory tests that allow us to develop hypotheses about what objects (sensor data) correspond to something known and observable by a human subject.
We use machine learning methods to classify three social scenarios from over a hundred experiments, demonstrating that it is possible to detect social situations with high accuracy, using the low-resolution sensors from our minimalist robot.
\end{abstract}

\section{Introduction}
In 2003, Rodney Brooks
suggested that ``by 2020 robots will be pervasive in our lives'' \citep[p. 113]{Brooks:2002aa}. As an example of this trend, he conceptualized an autonomous robotic vacuum cleaner for the home with a bottom-up design which allowed the robot's behaviors to emerge in interaction with its physical environment. The robot used the amount of light it sensed as a measure of the dirtiness of the floor, readings from its bump sensors as a signal to change direction, and its cliff sensors to know when to stop so as not to fall down stairs. Without full knowledge of the physical environment, the robot could randomly cover yet fully clean a wide variety of floors.  The \emph{iRobot Roomba} robotic vacuum, commercialized in 2002, is the materialization of Brooks' idea; more than 10 million Roombas have so far been sold worldwide.\footnote{http://www.irobot.com/en/us/Company/About.aspx?pageid=79}

As a robust, commercially available robotic product, the Roomba was one of the first robots to be used naturalistic and long-term studies of human-robot interaction in the home.  These pioneering studies ascertained that, along with the physical environment, the \emph{social context} also had an effect on the cleaning robot's ability to function successfully.  Researchers described that domestic Roombas were given names \citep{sung} and treated as ``social agents'' \citep{forlizzi}; the use of Roombas also had a reciprocal effect on the social organization and practices in the home, as men and teenagers participated more in domestic cleaning chores.  Such findings call attention to the importance of understanding the social as well as the physical dynamics of the context of use for robotic products.

Inspired by the Roomba as a commercial and social product, we propose that future robotic technologies that can co-exist and collaborate with people in everyday environments should have a sense of the social as well as physical contexts in which they operate. Contemporary robots are largely ignorant of the social significance of their actions and of the bustle of human life around them. As robots spend more time around humans, they will profit from being able to take advantage of the social as well as the physical characteristics of the environment to support their successful functioning \citep{Dautenhahn}.

Artificial Life has contributed greatly to the development of situated robots whose behavior emerges from the nonlinear interaction between machine and environment \citep{AlmeidaeCosta_Rocha:2005}.
But just like human cognition and social intelligence is extended into the environment \citep{clark1998being}, robots can use the bottom-up principles of artificial life to develop social competency.
Uexkull's concept of \emph{umwelt} (the self-centered sensorial world of animals) has served as a guiding principle to generate robot behavior that is grounded on their own perception-action interaction with an environment \citep{hoffmeyer1997signs}.
While the concept of an \emph{umwelt} makes the case for personal sensory experience, \cite{uexkull2001} notes that it gives us a way to understand sociality as an \emph{intersubjective} process rather than a subject-object dualism:
\begin{quote}
 \ldots the idea of an objective universe, that embraces all living things, is undeniably very useful for ordinary life. The conventional universe, where all our relationships to our fellow human beings are enacted, has brought all personal Umwelt spaces under a common denominator, and this has become indispensable for civilized human beings. Without it, we cannot draw the simplest map, because it is impossible to combine all subjective points of view in a single picture (p. 109).  
\end{quote}
Similarly, we now need to develop social umwelten for robots. By this we mean that, rather than designing robotic social behavior in a top-down manner, we need to develop it from the bottom-up in a manner that is most consistent with the robots own sensors and its interaction with the social environment (including designers).
This also means that the robot and humans must have more ways of recognizing, remembering, and building upon intersubjective experiences.
Within this multi-agency environment the idea is to develop umwelt overlap between robotic and human agents that scaffolds cooperative action \citep{Ferreira2013}.

We describe an initial approach to developing social awareness and presence for a domestic robot using minimalist robots, a combination of simple sensors, and bio-inspired computational techniques to develop a social umwelt for domestic robots.
We seek to develop socially situated robotic agents whose behavioral repertoire is subsumed by the social order of the space.
The goal is to recognize different classes of human activity and then associate to each class a range of appropriate behaviors. 
While human-robot interaction research has largely focused on developing algorithms that use high resolution data such as audio and video, our system relies on long time-series of low resolution data such as mechanical vibration, temperature, lighting, sounds and collisions.
We rely on low-resolution data because that is the reality of the sensors in the robot platform we use (see below). This allows us precisely to test if such cheap sensors, which are immediately and widely available, are capable of developing minimal social awareness.

We begin the paper with a discussion of the Roomba and its relationship to the social and cultural models of the home.
We then introduce the robot and test how it experiences the world through its sensors.  These exploratory tests allow us to develop hypotheses about what objects (sensor data) correspond to something known and observable by a human subject.
In the third section we use machine learning methods to classify three scenarios from over a hundred experiments involving human interaction. We conclude with some future directions for our research.

\section{Navigating social spaces}

The Roomba is one of the first instantiations of robots that work in everyday human environments with untrained users. Technology corporations and governments around the world expect that such technologies will proliferate and provide a new era of technological and economic production.
This future direction for robot development is highlighted by the US National Science Foundation's National Robotics Initiative, which funds the development of co-robots that ``work beside, or cooperatively with, people\ldots acting in direct support of and in a symbiotic relationship with human partners'' \citep{NRI2013}. We use the Roomba as a model case for studying how robotic technologies might be socially integrated into human environments.

Roomba's design, inspired by Brooks' subsumption approach to artificial intelligence, provides a robust and workable solution to issues posed by diverse and constantly changing human environments. Studies of Roomba's use in actual homes, however, have pointed out that there are important challenges and resources in the environment that the Roomba's design does not take into account.
For example, while the Roomba is designed to function in rooms of different shape, size, and organization, it is limited in the kinds of terrain it can cover. Therefore owners need to adapt the home to their Roombas by moving furniture, objects, and moving them between different levels of a house \citep{forlizzi}.
More relevant to our goals, Roomba's current limitations in terms of social awareness also limit its use; users may find the Roomba's random coverage of space may disrupt their activities, and therefore turn the Roomba on only when they are away from home or at specific times.

Existing research on robots in the home has shown that even functional robots like the Roomba are interpreted in social ways when they are situated in social environments  \citep{forlizzi}. The robot's emergent interactions with the social organization, cultural norms, political dynamics, and people's interpretations within the social environment can have a significant effect on whether the robot is accepted or rejected by users \citep{mutlu}. We have also shown that users envision and evaluate robots and other technologies in the context of the social hierarchies and relationships they regularly inhabit \citep{heerin}.  Initial research on service robots further suggests that personalization is an important component of robotic functions in open-ended human environments \citep{forlizzidisalvo}

The idea that human interpretations of a robot's functioning can support its behavioral repertoire was suggested decades ago in \cite{braitenberg1984} description of robotic ``vehicles'' whose simple behaviors in relation to the environment evoke ascriptions of affective and cognitive meaning by human observers.
Several social robotic projects, such as Keepon \citep{Kozima}, Muu \citep{Matsumoto}, and PARO \citep{Shibata}, have embraced the possibility of communicating social presence and agency through simple relational cues. \cite{alac} has shown that the social agency of robots is not constructed solely, or even primarily, through their functional capabilities, but is scaffolded through the ways in which human actors orient themselves towards the robot.

This research in human-robot interaction corroborates prior theories in human-computer interaction that describe everyday contexts as not only physical, but social and cultural spaces that hold both personal and shared meanings for their human inhabitants, which are constructed through embodied interaction \citep{dourish2001}.
We extend this understanding of how space relates to the design of robots and human-robot interaction by  developing minimalist robots that are becoming aware of the social and cultural characteristics of their environment.

\begin{figure}[t]
\begin{center}
\includegraphics[width=3.25in,angle=0]{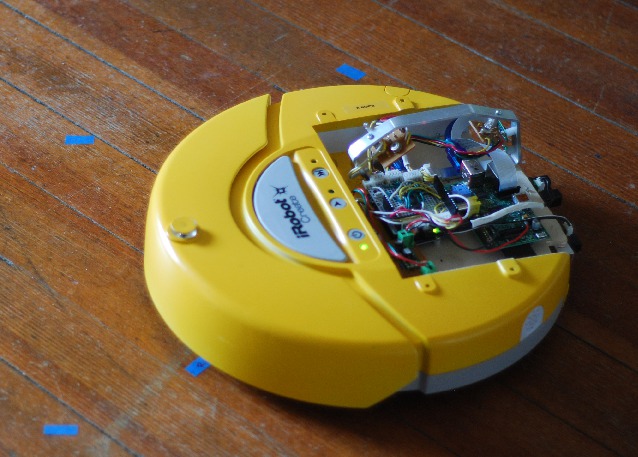}
\caption{Katie in operation.}
\label{yellow_katie}
\end{center}
\end{figure}

\section{The Katie family social robot}
In this section we give a description of our robots and the kind of sensorial world in which they operate.  The graphs of the sensor data in this section are meant to demonstrate the flow of data from the robots that can be interpreted by most people.  Can the robot tell the difference if it is next to a wall versus in the middle of a room?  If a person greets the robot, which sensors exhibit change and how much?  
The graphs of this data are a resource of the robot's social umwelt as it is the first place where both humans (researchers) and our robots establish a common denominator together with an object in the world.  Most importantly, graphs such as these help us to understand the limits of real time sensor responses, graphic visualization of data, and humans linking those to meaningful environmental changes.   To translate further between robot and human will require more sophisticated techniques, which we discuss in the next section.

\subsection{Robot design}

\begin{figure}[t]
\begin{center}
\includegraphics[width=2.4in,angle=0]{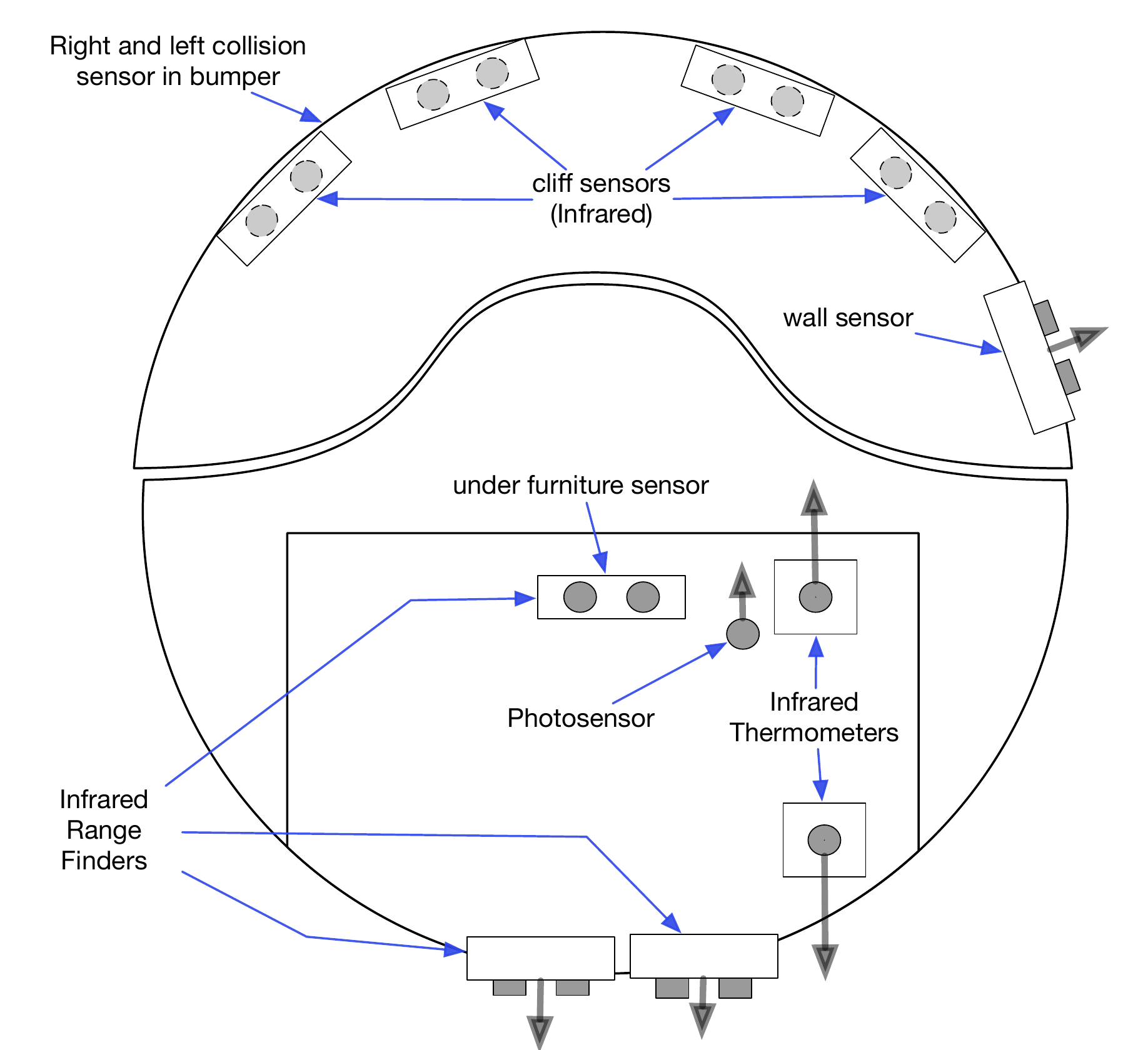}
\caption{Sensor locations on a Katie robot.}
\label{sensor_positions}
\end{center}
\end{figure}

The robots we are using, called the {\em Katie family}, are built on the {\em iRobot Create} platform, an {\em Arduino Uno}, and a {\em Raspberry Pi}.  The Create has several native sensors and we have included more sensors to the Arduino (see Figure \ref{sensor_positions}).  The Raspberry Pi collects images which are used for coding data and verifying our classes.   It also relays data to and commands from our server.  Our research group can begin and start experiments remotely and download data from each experiment.  These basic parts comprise the embodiment of the family and gives each their unique umwelt.

We chose this minimalist design for Katie to keep all components lightweight, low power, and housed inside the cargo bay of the Create.  This allows the robot to reside in a location for longer time spans and to explore (or seek shelter) under objects such as chairs, couches and tables. Retaining size and weight of the Create also maximizes mobility of the platform in physically tight or socially constrained indoor spaces, allowing the robot to do things that humans don't normally do in a space, such as getting on the ground to look under objects or to look closely at the ground and baseboards of a space.

\subsection{The sensorial world of Katie}

The results from multiple hour tests demonstrate the resolution of the sensor data, behavior of the sensors, and the starting range of patterns that comprise the robot's umwelt.  For designers, tests such as these help to decide where to place sensors and when to focus on a given data stream or not.  Because there is so much complexity in even some of our simplest spaces we give the robot a basic scanning behavior, which is a 30 degree pivot every 20 seconds.  

For all of our tests in this section and experiments in the next section, the following sensors are examined: two infrared (IR) range sensor aimed rearwards (integer)\footnote{The Sharp GP2Y0A02YK0F has a range of 15 to 150 cm and Sharp GP2Y0A41SK0F has a range of 4 to 30cm.}; a photo-voltaic cell for sensing light (integer); two IR thermometers (floats)\footnote{This sensor, the Melexis MLX90614, report hundredth's of a degree resolution, with an accuracy of $\pm .5 ^{\circ}\mathrm{C}$ for most ranges in room temperature and $\pm .1 ^{\circ}\mathrm{C}$ in human body temperature range.};  and Create platform internal sensors, namely 4 integer IR sensors around the bumper to detect cliffs, a wall IR sensor, and 5 boolean wheel and bumper sensors. This results in a total of 15 sensor variables.
An observation is recorded approximately once every tenth of a second.

\begin{figure}[t]
\begin{center}
\includegraphics[width=3.25in,angle=0]{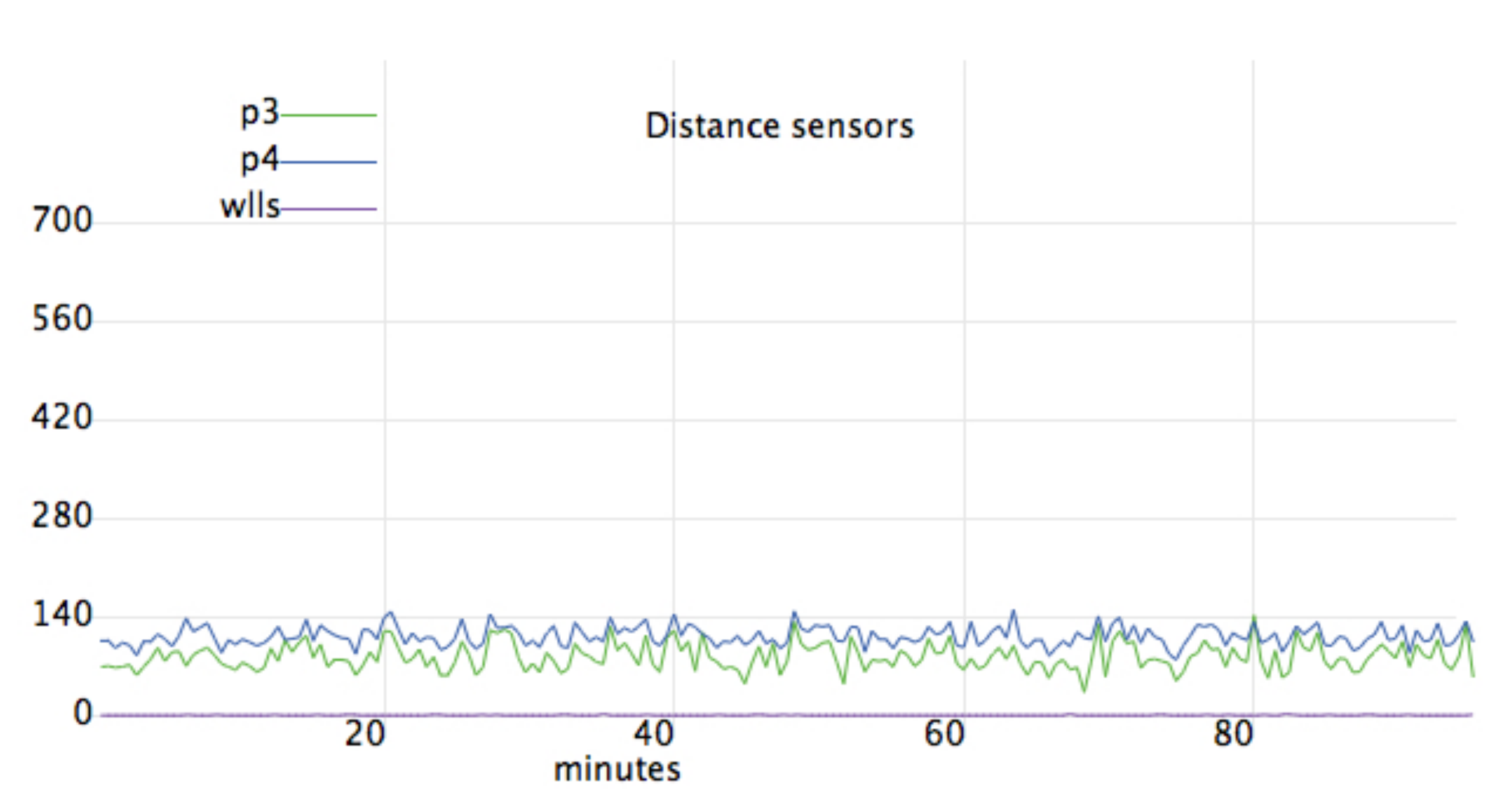}
\includegraphics[width=3.25in,angle=0]{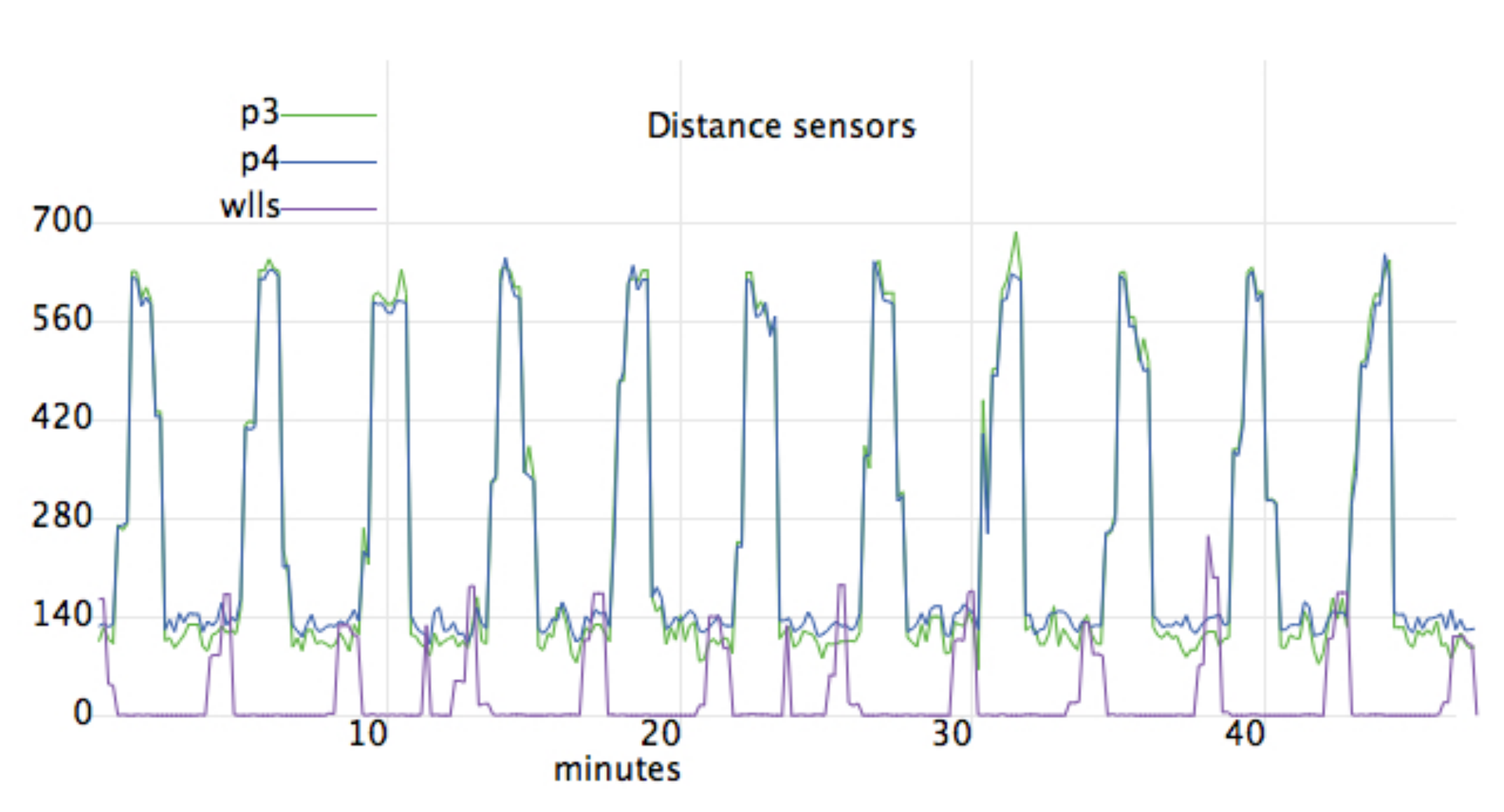}
\caption{Infrared distance sensor data from two positions.  The medium and long range sensors ($p3$ and $p4$) disagree little.  In future tests the long range sensor $p4$ will face upward to detect when the robot is under furniture and possibly detect when in a door frame. }
\label{fig1}
\end{center}
\end{figure}

The robots have multiple infrared distance sensors.  There are three outward facing sensors on the robot each with a different range.   Figure \ref{fig1} shows two 90 minute runs of data collected from a robot positioned in an open floor with no objects nearby and one positioned next to a wall.
On the Create there are an additional four distance sensors that are pointed toward the ground for cliff detection.   The short range on these allow for detection of small position changes if, for instance, the robot is moved or if the floor moves.  

\begin{figure}[t]
\begin{center}
\includegraphics[width=3.25in,angle=0]{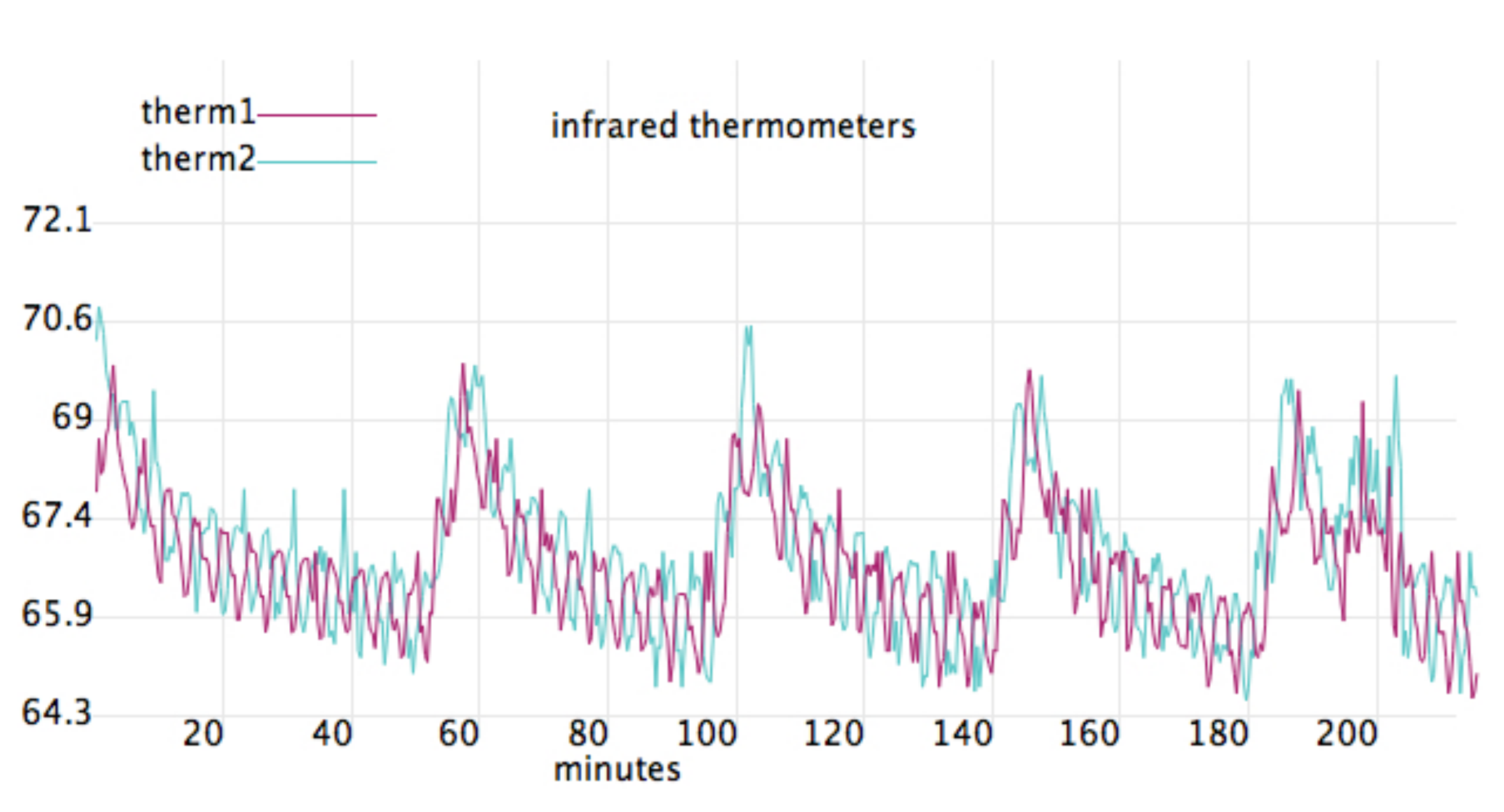}
\caption{Infrared thermometers pick up changes as the robot faces a new direction every 20 seconds and when the furnace turns on in the house.}
\label{fig2}
\end{center}
\end{figure}

There are two infrared thermometers positioned on the robot at an upward angle of 45 degrees.  Each face in opposite directions.  Data from a four hour test in Figure \ref{fig2} shows two cyclical patterns.  The first cycle is from the rotation of robot in its scan behavior where we see a $1 ^{\circ}\mathrm{F}$ difference in the sensors at one point in each rotation. 

\begin{figure}[t]
\begin{center}
\includegraphics[width=2.5in,angle=0]{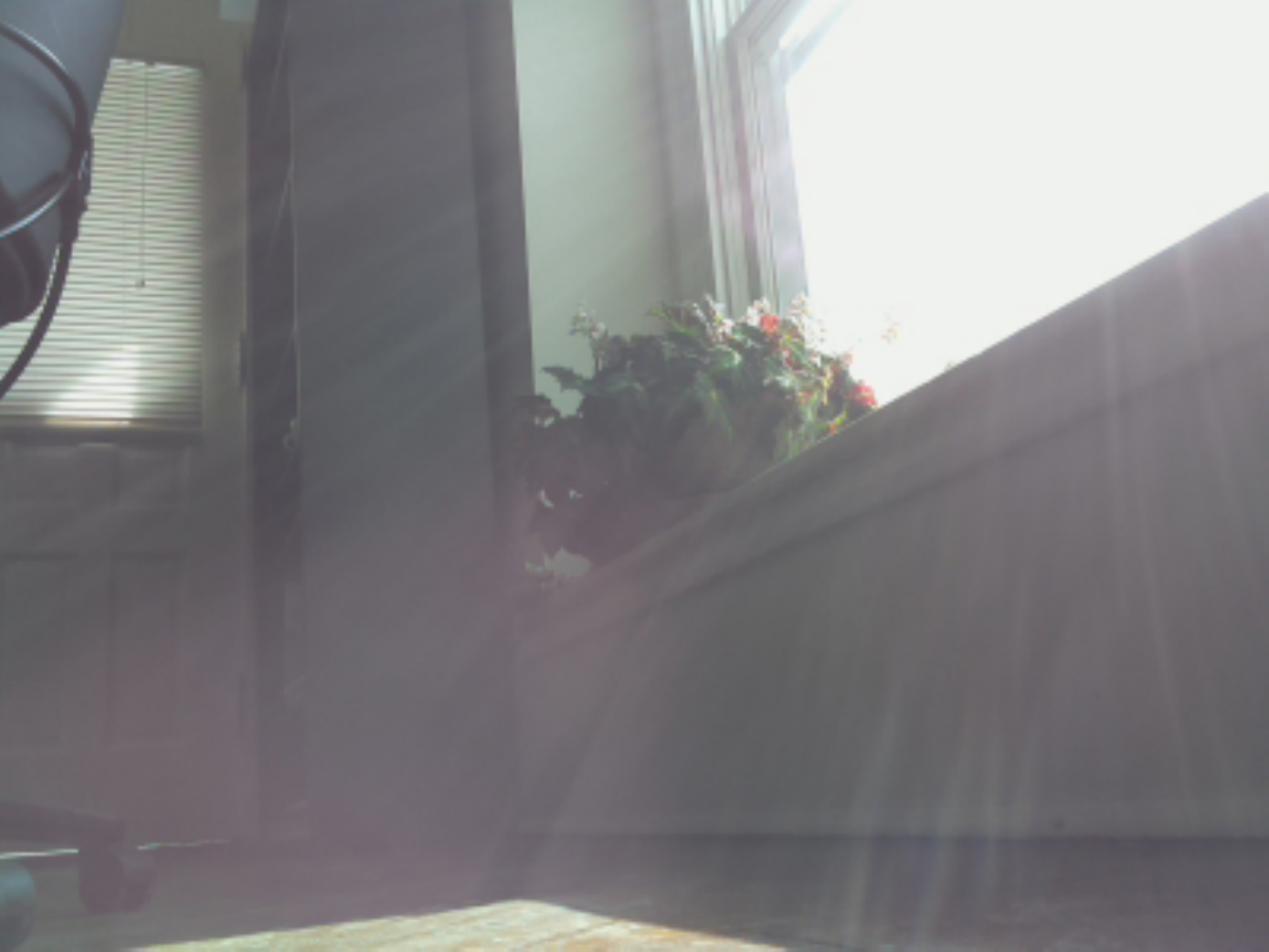}
\includegraphics[width=3.25in,angle=0]{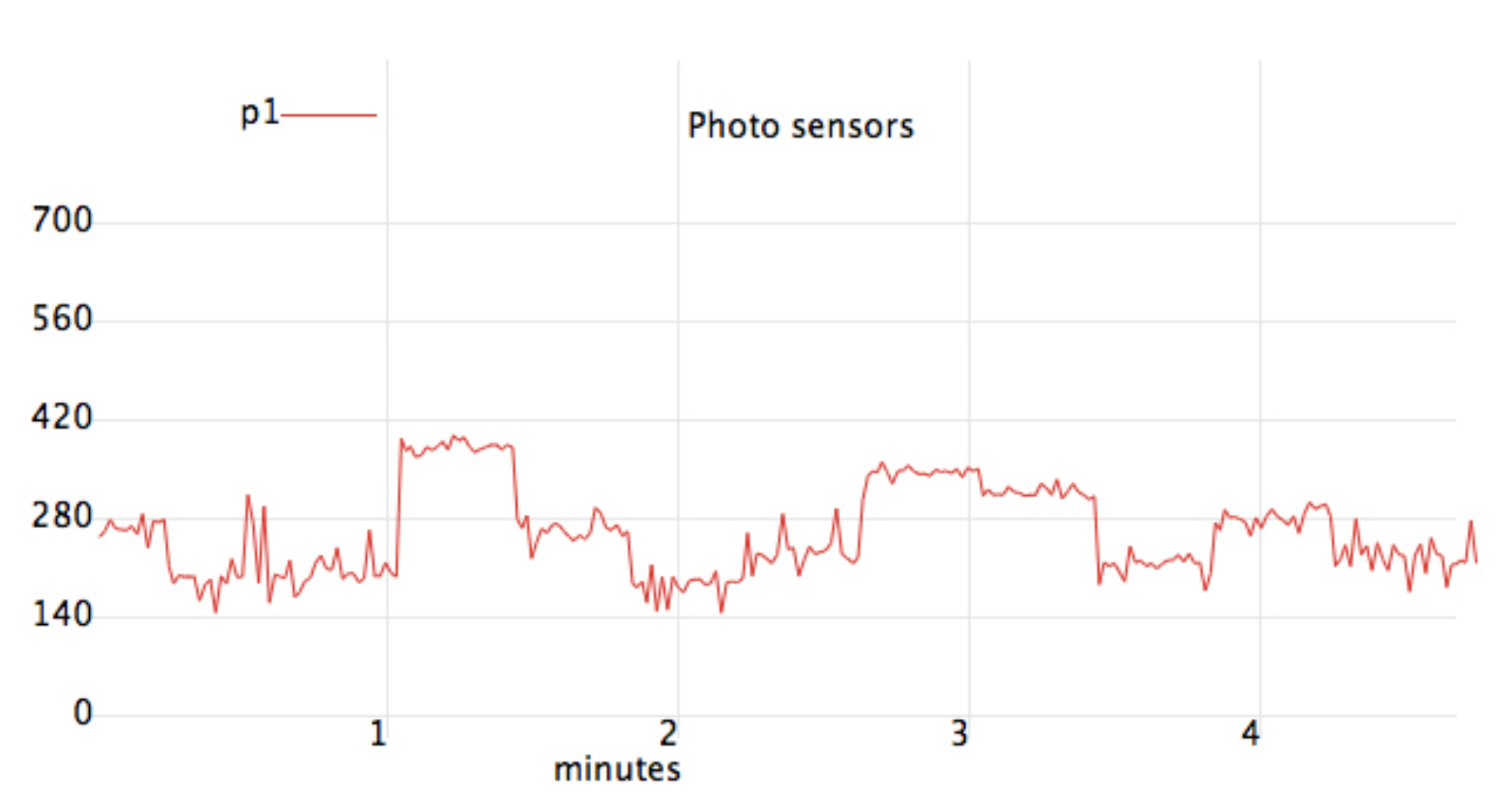}
\caption{Light can provide some directional context if the photo sensor has a slight angle and the light source is intense.  The picture was taken from the robot while it was collecting data in the graph.}
\label{light}
\end{center}
\end{figure}

The light sensor has a slight angle towards the front of the robot.  An angle on this sensor give some direction of where light is coming into a room.  Figure \ref{light} shows data when the robot is in a sunny location in a house.  Depending on the intensity of light in a specific direction, fast changes in the light can be caused by humans or animals casting a shadow on the sensor.  Very fast changes, like an activation of an electric lamp, cause sharp and consistent patterns from the light sensor reading while changes in the light coming through windows caused by clouds are smooth and stable.

While the robot is stationary, the bump sensors are our the most reliable detector of presence of people and animals.  The bump sensor is a large plastic bumper that covers the entire front end of the robot with a switch on each side.  When an object depresses the right side the right switch is triggered.  Contact with the center of the bumper triggers both sensors.  All three states (contact with left side is $bump = 1$, right side is $bump = 2$, and center is $bump = 3$).  The wheel drop sensors are also switches.   These trigger when the wheels extend all the way down (each wheel has spring to force them down when they are off the floor) are are therefore reliable for detecting when the robot is lifted.

Finally each Katie has a high resolution camera.  The main purpose of the camera is to annotate the data with socially meaningful categories and to verify if the robot's classifications make sense.   In the current form the camera is angled slightly upward.  Robot mounted cameras can collect potentially sensitive or embarrassing information and this increases within a private space like a home. 
Since Americans are becoming increasingly aware and concerned about privacy we design into the robots some deference to privacy by mounting the camera with an angle toward the ground.  Lowering the gaze is an embodied signal that shows deference to broader cultural concerns.  
The images offer another affordance that will be crucial as the robot learns more about the household and begins to move easily within it.  Images can be used by people in the home to identify meaningful objects and places and such annotations can used to respond to a richer understanding of the environment.

\section{Classifying social situations}

\begin{figure}[t]
\centering
\includegraphics[width=3.4in]{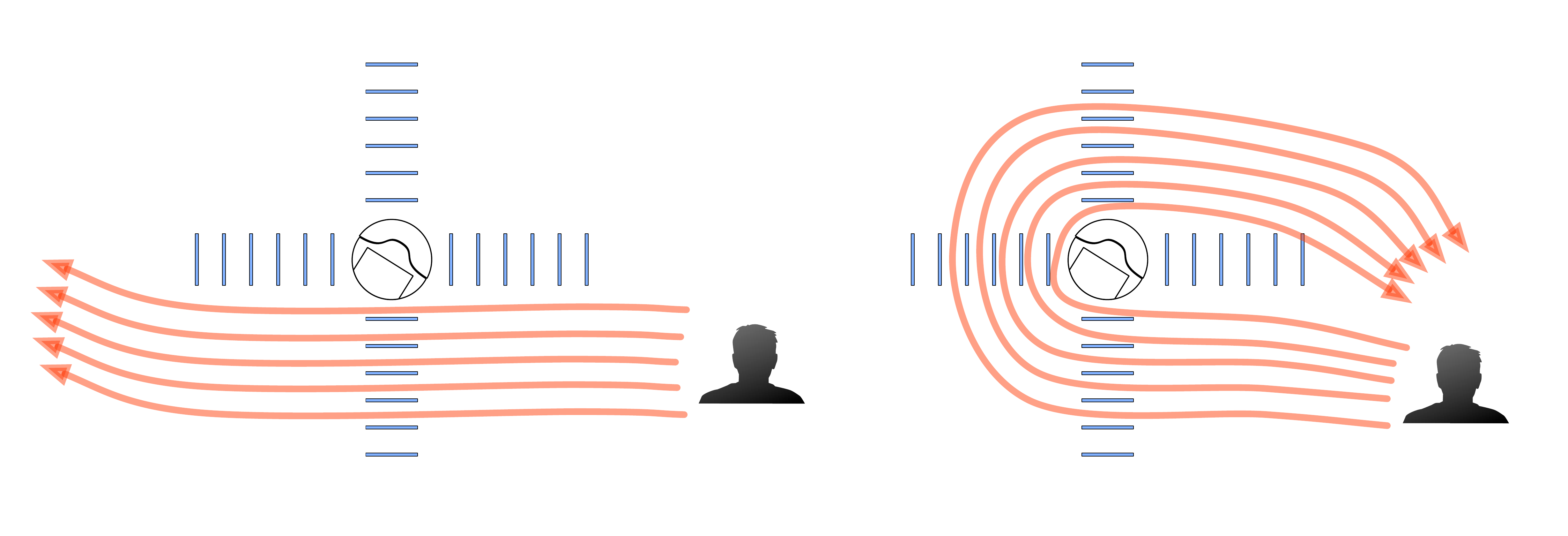}
\caption{Scenarios 1 (left) and 2 (right).}
\label{scenarios}
\end{figure}

As a first step in developing the social capabilities of the system, we would like to know if a Katie can detect differences in simple human-robot interactions.
The scenarios we wish to discriminate between are: (0) an empty room, (1) someone walking across a room and (2) someone walking around the robot.
Scenarios 1 and 2 are represented in figure \ref{scenarios}. The robot is placed in the center of a room with measurements of distance from the robot taped to the floor. Each scenario is run at 5 different proximities: contact with the robot, 1-20 cm, 21-40 cm, 41-60 cm, and 61-80cm.
Each scenario-proximity condition is run 10 times, for a total of 150 labeled \emph{experiments}. For each experiment, two doors were randomly selected (by computer) from the three entrances to the room to be the starting and ending doors.
During all experiments, the robot performed its scanning behavior, turning 30 degrees every 20 seconds. Certain sensors like the photo-sensor and the thermometers are correlated with time due to natural ambient variations. In order to create independence between the scenarios and the observations, the order in which the 150 experiments were conducted was random. The robot recorded an \emph{observation}, a set of readings from its sensors, about once every 0.1 to 0.2 seconds.
The mean (and standard deviation) of the number of observations per scenarios 0, 1, and 2 are: 134.92 (57.18), 38.54 (15.59) and 70.46 (21.19), respectively.

We are also interested in understanding how important various sensors are to the performance of the system, in order to avoid building new versions of the robot with useless sensors, and to help decide what new sensors should be added to the robot.

\subsection{Methods}

To classify the labeled data we used three well-known classifiers implemented in the Python sckit-learn library \citep{scikit-learn}: \emph{random forest}, \emph{boosting}, and \emph{logistic regression}. The first two are decision-tree classifiers, the last is a maximum-likelihood method that separates data based on linear relationships between variables.
Ten-fold cross-validation was performed for each classifier as follows:
the validation set of each fold contains a single, randomly-selected experiment from every scenario-proximity condition, for a total of 15 out of 150 experiments; the remaining 135 experiments (9 from each condition) comprise the training set of each fold.
Sensor data is normalized by subtracting its mean and standard deviation calculated from the training set, which is especially helpful for logistic regression.

Decision tree learning is useful for classifying data with nonlinear relations. A decision tree partitions the data into regions through recursive binary splits, choosing the best predictor for the split at each step according to an impurity measure.
We choose the \emph{Gini index} for training, since it is more sensitive than misclassification error and more interpretable than cross-entropy. This impurity measure can be interpreted as the training error rate at the split.
However, to evaluate the performance on validation data, we use standard \emph{misclassification error}
\citep{hastie2009elements}.

The growth of decision trees is highly sensitive to noise in the data. Any errors in the first splits are propagated down to all splits below it.
To reduce this variance, ensemble methods like bagging, boosting, and random forest can be used. These ensemble methods produce a forest of trees, with the final classification determined by a majority vote among them \citep{hastie2009elements}. We use two such ensemble methods: \emph{random forest}, which creates trees trained on bootstrapped data with limited access to variables at each split; and the \emph{SAMME boosting} algorithm, which iteratively trains trees while weighting data points by their difficulty of classification, and weighting trees by their training accuracy.
These methods should be able to perform well even in the presence of highly nonlinear patterns in the data, and deal well with ambiguous data points.

Finally, \emph{logistic regression} classifies data according to linear relationships between predictors. The coefficients of this relationship are estimated by maximizing the log-likelihood of the conditional probability distribution of the class given an observation, modeled as a linear function of the variables. This classifier should do well if the data can be linearly separated, and has the additional benefit of returning a probability of class membership for future analysis.
Two regularization penalties to the size of the regression coefficients are investigated: L1 (linear) and L2 (squared). These penalties subtract from the objective function of the regression, respectively, the sum of the absolute values of the coefficients and the sum of the squares of the coefficients.
When the data are normalized to similar ranges, this can be used for variable selection \citep{hastie2009elements}.

\subsection{Results}

\begin{figure}[t]
	\centering
	\includegraphics[width=3.25in]{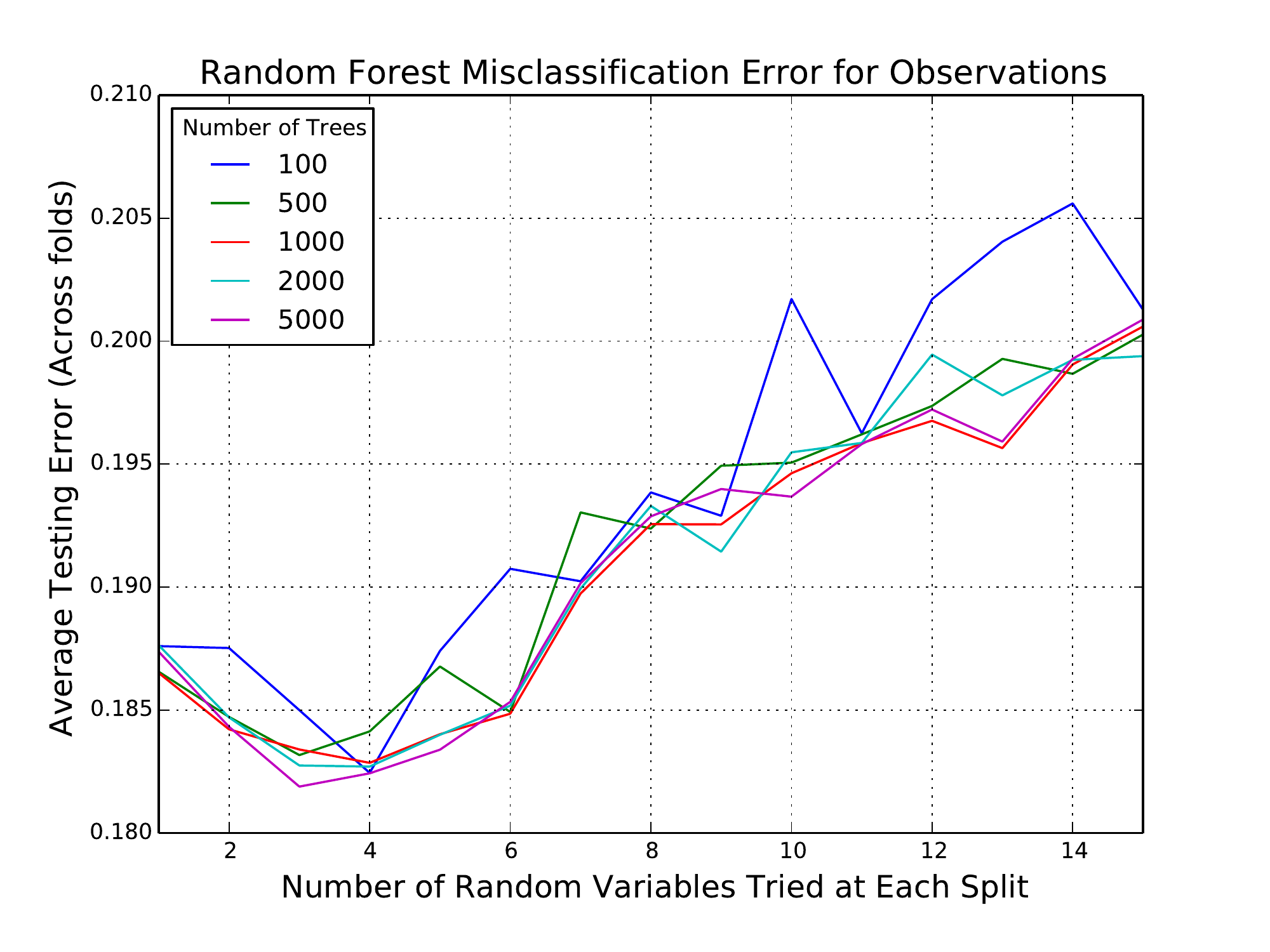}
	\caption{Mean cross-validation error of Random Forest classifier over observations, for number of trees and number of variables at each split.}
	\label{randomForestPerObservation}
\end{figure}

\begin{figure}[t]
	\centering
	\includegraphics[width=3.25in]{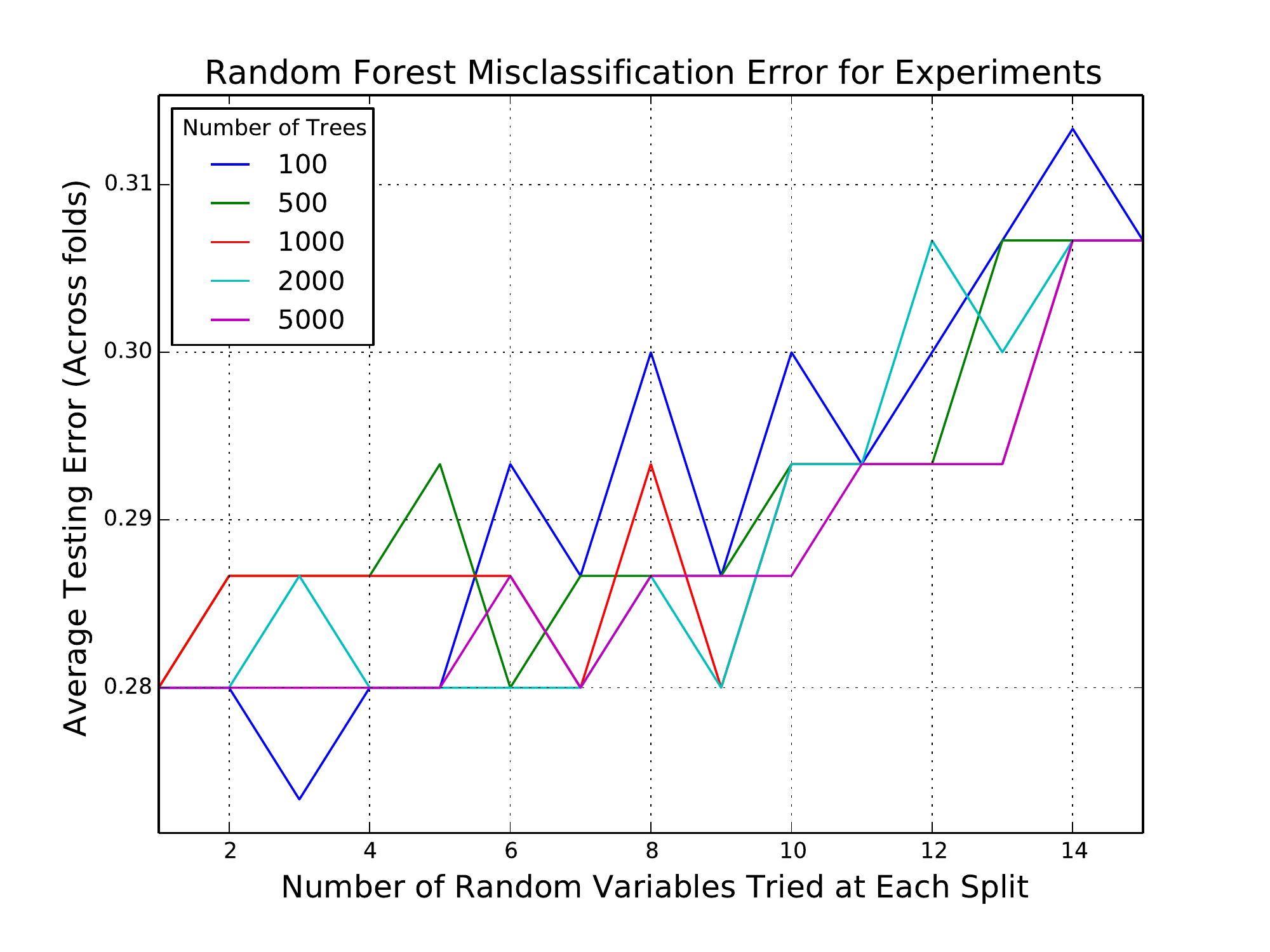}
	\caption{Mean cross-validation error of Random Forest classifier over experiments, for number of trees and number of variables at each split.}
	\label{randomForestPerExperiment}
\end{figure}

We analyzed classifier performance for both observations and experiments.
The average validation misclassification error of the random forest across all ten cross-validation folds is depicted in figures \ref{randomForestPerObservation} and \ref{randomForestPerExperiment} for observations and experiments, respectively.
Since we ultimately want the robot to classify scenario-proximity conditions, rather than single sensor observations, each experiment is classified according to a majority vote among the labels predicted for observations taken during that experiment.
This also weighs each scenario-proximity condition equally, whereas per-observation error favors conditions with the most data.
The performance is fairly robust to the number of trees, although more trees, as expected, tends to produce a smoother curve (more robust to changes in number of variables tried at each tree split).
We can see that performance tends to be best when the classifier has access to 3 or 4 randomly selected variables at each split.

\begin{table}
	\centering
    \begin{tabular}{ | l | l | l |}
    	\hline
    	Classifier & Obs. Error & Exp. Error \\ \hline \hline
    	Random Forest & $0.182 \pm .041$ & $0.273\pm .094$ \\ \hline
    	SAMME Boosting & $0.204 \pm .055$ & $0.293\pm .106$ \\ \hline
    	Log. Reg. & $0.209 \pm .065$ & $0.320 \pm .086$ \\ \hline
    	Trivial & $0.711 \pm .019$ & $0.667 \pm .000$\\ \hline
    	Random & $0.586 \pm .007$ & $0.664 \pm .001$\\
    	\hline
    \end{tabular}
    \caption{Observation and experiment errors on 3-Scenario classification. Mean and 95\% confidence interval.}
    \label{classifierError}
\end{table}

\begin{figure}[t]
	\centering
	\includegraphics[width=3.25in]{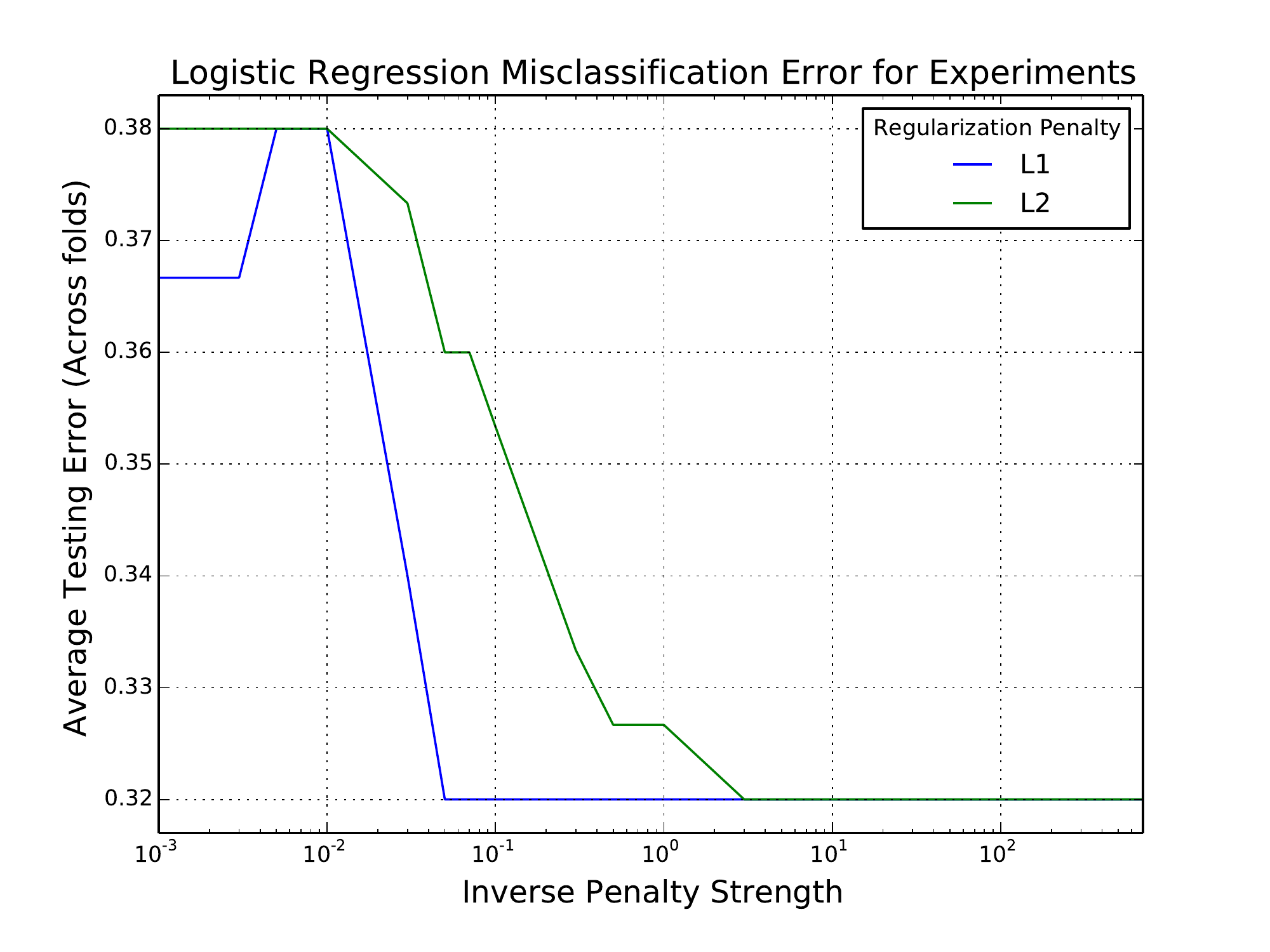}
	\caption{Mean cross-validation error of Logistic Regression classifier over experiments, for both regularization penalties}
	\label{lrPerExperiment}
\end{figure}

The best classifier parameters were selected according to the average validation misclassification error across all ten cross-validation folds.
The corresponding errors and their 95\% confidence interval are shown in Table \ref{classifierError}. The confidence intervals were calculated using the errors on each fold, assuming a t-distribution.
The random forest classifier achieves observation and experiment mean error rates as low as 0.182 and 0.273, respectively.
The boosting classifier
achieved similar performance (figures with mean cross-validation error not shown).
The performance of the logistic regression classifiers per experiment is shown in figure \ref{lrPerExperiment}. Performance improves with smaller regularization penalties, and for sufficiently small penalties, the performance is not significantly worse than random forest.
Overall, the three classifiers can classify correctly about 80\% of the time in which of the three scenarios an observation was taken, and the experiments 70\% of the time.

We also computed the performance of two null-model classifiers. The \emph{trivial} classifier labels every observation in the test set with the most frequent class label in the training set of each fold. The \emph{random} classifier randomly labels observations in the test set with the same frequency that classes appear in the training set.
The three (non-null) classifiers perform quite well, given the low-resolution sensors. They significantly outperformed the null models, although not significantly different from each other.
This suggests that any nonlinear patterns in the data are not significant for performance on this minimal social classification task---since logistic regression was not significantly worse than decision-tree classifiers.

While the performance of the classifiers is significantly better than null models, it could still be the case that most of the error would be between the two walking scenarios 1 and 2.
These scenarios may be harder to distinguish, since both begin the same way. %
It is important that a social robot be able to determine an empty room from an occupied one.
However, it is more important for social awareness that Katie can distinguish between an indifferent person and one that displays an interest in interacting, situations whose proxies here are scenarios 1 and 2 respectively.

In order to investigate whether these two scenarios are discriminated well by the same classifiers, we performed binary classifications, where scenario 2 is ``positive'' and scenario 1 is ``negative''.
These values can be interpreted as answers to the question: does a person want to interact with the Katie?
The average ten-fold cross-validation performance of the classifiers is presented in Tables \ref{binaryStats} and \ref{binaryVoteStats} for the best parameters. Performance is reported for accuracy, balanced F1 measure (harmonic mean of precision and sensitivity) and the Matthew's correlation coefficient (MCC)\footnote{\label{emptybox} The trivial classifier for observations and experiments, and the random classifier for experiments label all data into a single class, which results in a division by 0 in the calculations of the MCC}.

\begin{table}
	\centering
    \begin{tabular}{ | l | l | l | l |}
    	\hline
    	Classifier & Accuracy & F1 & MCC \\ \hline \hline
    	R. Forest 	& $0.874 \pm .077$ & $0.893 \pm .079$ & $0.765 \pm .118$ \\ \hline
    	SAMME 			& $0.852 \pm .057$ & $0.882 \pm .053$ & $0.703 \pm .104$ \\ \hline
    	Log. Reg. 		& $0.845 \pm .078$ & $0.861 \pm .083$ & $0.734 \pm .112$ \\ \hline
    	Trivial 		& $0.647 \pm .029$ & $0.785 \pm .021$ & --- \footnotemark[\getrefnumber{emptybox}] \\ \hline
    	Random 			& $0.542 \pm .000$ & $0.645 \pm .000$ & $-0.001 \pm .001$ \\
    	\hline
    \end{tabular}
    \caption{2-Scenario classification for observations}
    \label{binaryStats}
\end{table}
\begin{table}
	\centering
    \begin{tabular}{ | l | l | l | l |}
    	\hline
    	Classifier & Accuracy & F1 & MCC \\ \hline \hline
    	R. Forest & $0.900 \pm .075$ & $0.892\pm .094$ & $0.826 \pm .128$\\ \hline
    	SAMME & $0.860 \pm .077$ & $0.856\pm .092$ & $0.748 \pm .139$ \\ \hline
    	Log. Reg. & $0.880 \pm .066$ & $0.862 \pm .090$ & $0.789 \pm .110$\\ \hline
    	Trivial & $0.500 \pm 0$ & $0.667 \pm .000$ & --- \footnotemark[\getrefnumber{emptybox}] \\ \hline
    	Random & $0.519 \pm .001$ & $0.673 \pm .001$ & --- \footnotemark[\getrefnumber{emptybox}] \\
    	\hline
    \end{tabular}
    \caption{2-Scenario performance for experiments}
    \label{binaryVoteStats}
\end{table}

The performance of the classifiers is again quite good and significantly better than the null models, though not significantly different from one another.
It is clear that the two social scenarios can be distinguished by Katie's low-resolution sensors most of the time; with accuracy reaching 90\% of the time with random forest.
The MCC measures the correlation between observed and predicted labels. It is zero for random prediction (as is the case of our random null model).
For experiments, it reaches 0.826 which is a very high correlation between observation and prediction.
It is also worth noticing that in the 2-scenario classification, contrary to the 3-scenario case, the performance was slightly higher for experiments than observations. This is likely due the larger number of observations gathered by scenario 0 experiments.

\begin{figure}[t]
	\centering
	\includegraphics[width=3.25in]{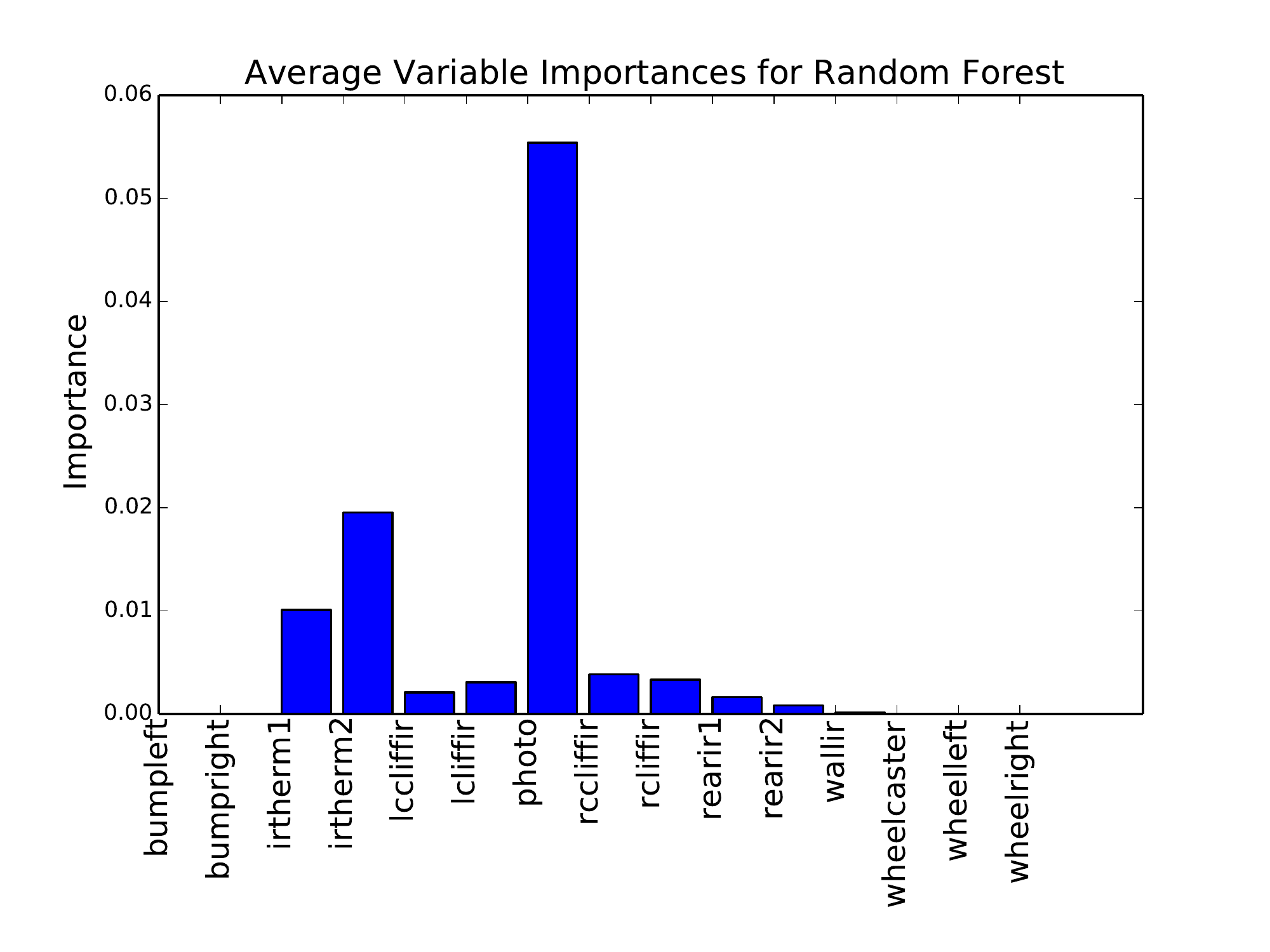}
	\caption{Random Forest variable importance}
	\label{randomForestImportances}
\end{figure}

The relative importance of each sensor to classification performance is calculated as the expected fraction of observations that each sensor variable contributes to in the classification.
This is depicted in figure \ref{randomForestImportances} for the random forest classifiers.
In this case, the photo sensor is the most useful, followed by the IR thermometers, the cliff sensors, and the rear-facing IR range sensors. The bumps, wheel, and wall sensor are not useful for discriminating these social scenarios.
Results are similar for the boosting classifier, but with the photo sensor greatly emphasized (figure not shown).

As a proxy for variable importance, in the case of logistic regression, we can examine the average coefficients produced by the classifier for variables across folds (figure not shown). However, different sets of coefficients correspond to different scenarios. As in the case of the random forest and boosting classifiers, the photo sensor and IR thermometers are found to be important for all scenarios, but the bump sensors are also relevant in distinguishing scenario 0, since they are not activated during any scenario 0 experiment.

\section{Future work}

We see robotic development as moving from an understanding of robots as techno-scientific artifacts to sociotechnical ones.
The social umwelt is woven together through technical infrastructure and we have illustrated this infrastructure with a simple robot.
Furthermore, we showed that such a robot endowed with low-resolution sensors is capable of distinguishing minimal social scenarios with high accuracy.
We plan on designing more sophisticated communication loops  for humans and robots to come to a shared experience of their environment.  These will extend beyond the social world of our household lab and incorporate ways for the robot to select different forms of motion.
To move in this direction of more complicated social robot umwelts, we will design robots capable of interacting directly with people running common household scenarios such as arriving home from work, reading a book, watching tv, etc.   We will use this information for further classification tasks and hope to refine the kinds of social states the robot can detect using simple sensors.  At the same time we will have people interact with the backend of the system, looking at the world from the perspective of the robot, showing graphs and associated images, and classifying sensor data.   Once the robot has a basic repertoire of meaningful classifiers it can begin making guesses about interesting/anomalous social events and then eliciting humans for annotations.

\section{Acknowledgements}

This work was supported by an Indiana University Collaborative Research Grant.

\footnotesize
\bibliographystyle{apalike}

\bibliography{ALife_Social_SLAM_FINAL_arXiv_submission}

\begin{thebibliography}{}

\bibitem[Ala\v{c} et~al., 2011]{alac}
Ala\v{c}, M., Movellan, J., and Tanaka, F. (2011).
\newblock When a robot is social: Spatial arrangements and multimodal semiotic
  engagement in the practice of social robotics.
\newblock {\em Social Studies of Science}, 41(6):893--926.

\bibitem[Almeida~e Costa and Rocha, 2005]{AlmeidaeCosta_Rocha:2005}
Almeida~e Costa, F. and Rocha, L.~M. (2005).
\newblock Embodied and situated cognition.
\newblock {\em Artif. Life}, 11(1-2):5--12.

\bibitem[Braitenberg, 1984]{braitenberg1984}
Braitenberg, V. (1984).
\newblock {\em Vehicles: Experiments in synthetic psychology}.
\newblock MIT Press, Cambridge, MA.

\bibitem[Brooks, 2002]{Brooks:2002aa}
Brooks, R.~A. (2002).
\newblock {\em Flesh and machines: How robots will change us}.
\newblock Pantheon Books, New York.

\bibitem[Clark, 1998]{clark1998being}
Clark, A. (1998).
\newblock {\em Being there: Putting brain, body, and world together again}.
\newblock MIT press.

\bibitem[Dautenhahn et~al., 2002]{Dautenhahn}
Dautenhahn, K., Ogden, B., and Quick, T. (2002).
\newblock From embodied to socially embedded agents - implications for
  interaction-aware robots.
\newblock {\em Cognitive Systems Research}, 3:397--428.

\bibitem[Dourish, 2001]{dourish2001}
Dourish, P. (2001).
\newblock {\em Where the action is: The foundations of embodied interaction}.
\newblock MIT Press.

\bibitem[Ferreira and Caldas, 2013]{Ferreira2013}
Ferreira, M. I.~A. and Caldas, M.~G. (2013).
\newblock The concept of {U}mwelt overlap and its application to cooperative
  action in multi-agent systems.
\newblock {\em Biosemiotics}, 6(3):497--514.

\bibitem[Forlizzi, 2007]{forlizzi}
Forlizzi, J. (2007).
\newblock How robotic products become social products: An ethnographic study of
  cleaning in the home.
\newblock In {\em ACM/IEEE International Conference on Human-Robot Interaction
  (HRI)}, volume~2, pages 129--137. ACM.

\bibitem[Forlizzi and DiSalvo, 2006]{forlizzidisalvo}
Forlizzi, J. and DiSalvo, C. (2006).
\newblock Service robots in the domestic environment: A study of the {R}oomba
  vacuum in the home.
\newblock In {\em ACM SIGCHI/SIGART Conference on Human-Robot Interaction
  (HRI)}, HRI '06, pages 258--265, New York, NY, USA. ACM.

\bibitem[Hastie et~al., 2009]{hastie2009elements}
Hastie, T., Tibshirani, R., Friedman, J., Hastie, T., Friedman, J., and
  Tibshirani, R. (2009).
\newblock {\em The elements of statistical learning}, volume~2.
\newblock Springer.

\bibitem[Hoffmeyer, 1997]{hoffmeyer1997signs}
Hoffmeyer, J. (1997).
\newblock {\em Signs of meaning in the universe}.
\newblock Indiana University Press.

\bibitem[Kozima et~al., 2009]{Kozima}
Kozima, H., Michalowski, M.~P., and Nakagawa, C. (2009).
\newblock Keepon: A playful robot for research, therapy, and entertainment.
\newblock {\em International Journal of Social Robotics}, 1(1):3--18.

\bibitem[Lee and Sabanovic, 2013]{heerin}
Lee, H.~R. and Sabanovic, S. (2013).
\newblock Culturally variable preferences for robot design and use in {S}outh
  {K}orea, {T}urkey, and the {U}nited {S}tates.
\newblock In {\em ACM/IEEE International Conference on Human-Robot Interaction
  (HRI)}, pages 17--24.

\bibitem[Matsumoto et~al., 2005]{Matsumoto}
Matsumoto, N., Fujii, H., Goan, M., and Okada, M. (2005).
\newblock Minimal design strategy for embodied communication agents.
\newblock In {\em IEEE International Workshop on Robot and Human Interactive
  Communication (RO-MAN)}, volume~14, pages 335--340.

\bibitem[Mutlu and Forlizzi, 2008]{mutlu}
Mutlu, B. and Forlizzi, J. (2008).
\newblock Robots in organizations: The role of workflow, social, and
  environmental factors in human-robot interaction.
\newblock In {\em ACM/IEEE International Conference on Human-Robot Interaction
  (HRI)}, volume~3, pages 287--294.

\bibitem[{National Science Foundation}, 2013]{NRI2013}
{National Science Foundation} (2013).
\newblock National robotics initiative.

\bibitem[Pedregosa et~al., 2011]{scikit-learn}
Pedregosa, F., Varoquaux, G., Gramfort, A., Michel, V., Thirion, B., Grisel,
  O., Blondel, M., Prettenhofer, P., Weiss, R., Dubourg, V., Vanderplas, J.,
  Passos, A., Cournapeau, D., Brucher, M., Perrot, M., and Duchesnay, E.
  (2011).
\newblock Scikit-learn: Machine learning in {P}ython.
\newblock {\em Journal of Machine Learning Research}, 12:2825--2830.

\bibitem[Shibata, 2012]{Shibata}
Shibata, T. (2012).
\newblock Therapeutic seal robot as biofeedback medical device: Qualitative and
  quantitative evaluations of robot therapy in dementia care.
\newblock {\em Proceedings of the IEEE}, 100(8):2527--2538.

\bibitem[Sung et~al., 2007]{sung}
Sung, J.-Y., Guo, L., Grinter, R.~E., and Christensen, H.~I. (2007).
\newblock ``{M}y roomba is rambo'': Intimate home appliances.
\newblock In {\em International conference on Ubiquitous computing (UbiComp)},
  volume~9 of {\em UbiComp}, pages 145--162. ACM.

\bibitem[Uexkull, 2001]{uexkull2001}
Uexkull, J. (2001).
\newblock An introduction to umwelt.
\newblock {\em Semiotica}, 134(1):107--110.

\end{thebibliography}

\end{document}